\documentclass[11pt,3p,review,authoryear]{elsarticle}
\let\today\relax
\makeatletter
\def\ps@pprintTitle{%
    \let\@oddhead\@empty
    \let\@evenhead\@empty
    \def\@oddfoot{\footnotesize\itshape
         {Preprint available on arXiv.} \hfill\today}%
    \let\@evenfoot\@oddfoot
    }
\usepackage{array}
\usepackage{amsmath} 
\usepackage{mathtools}
\usepackage{dsfont}
\usepackage{url}
\biboptions{longnamesfirst}
\usepackage[dvipsnames]{xcolor}
\usepackage{hhline}
\usepackage{appendix}
\usepackage{subfig}
\usepackage{soul}
\usepackage{tabularx}
\usepackage{threeparttablex}
\usepackage{booktabs}
\usepackage{array}
\newcolumntype{L}[1]{>{\raggedright\let\newline\\\arraybackslash\hspace{0pt}}p{#1}}
\newcolumntype{C}[1]{>{\centering\let\newline\\\arraybackslash\hspace{0pt}}p{#1}}
\newcolumntype{R}[1]{>{\raggedleft\let\newline\\\arraybackslash\hspace{0pt}}p{#1}}

\usepackage{makecell}
\usepackage{parskip}

\begin{document}
\begin{frontmatter}

\title{The impact of data set similarity and diversity on transfer learning success in time series forecasting}

\author[inst1,inst2]{Claudia~Ehrig\corref{cor}}
\author[inst1,inst3,inst4]{Benedikt~Sonnleitner}
\author[inst1]{Ursula~Neumann}
\author[inst2]{Catherine~Cleophas}
\author[inst5,inst6]{Germain~Forestier}
	
\affiliation[inst1]{
organization={Fraunhofer IIS, Fraunhofer Institute for Integrated Circuits IIS, Fraunhofer Center for Applied Research on Supply Chain Services SCS},
country={Germany}}
		
\affiliation[inst2]{organization={Christian-Albrechts-Universität zu Kiel, Research Group Service Analytics},
country={Germany}}

\affiliation[inst3]{organization={Otto-Friedrich-Universität Bamberg, Supply Chain Management},
country={Germany}}

\affiliation[inst4]{organization={Vives University of Applied Science, Business Management},
country={Belgium}}
		
\affiliation[inst5]{
organization={University of Haute-Alsace, IRIMAS}, 
country={France}}
       
\affiliation[inst6]{organization={Monash University, Department of Data Science and Artificial Intelligence},
country={Australia}} 

\cortext[cor]{Corresponding author: claudia.ehrig@iis.fraunhofer.de}


\begin{abstract}
Pre-trained models have become pivotal in enhancing the efficiency and accuracy of time series forecasting on target data sets by leveraging transfer learning. While benchmarks validate the performance of model generalization on various target data sets, there is no structured research providing similarity and diversity measures to explain which characteristics of source and target data lead to transfer learning success. 
Our study pioneers in systematically evaluating the impact of source-target similarity and source diversity on zero-shot and fine-tuned forecasting outcomes in terms of accuracy, bias, and uncertainty estimation. We investigate these dynamics using pre-trained neural networks across five public source datasets, applied to forecasting five target data sets, including real-world wholesales data. We identify two feature-based similarity and diversity measures, finding that source-target similarity reduces forecasting bias, while source diversity improves forecasting accuracy and uncertainty estimation, but increases the bias.
\end{abstract}


\end{frontmatter}

\section{Introduction}  
Transfer learning for time series forecasting describes pre-training a model on one or more source data sets in order to either continue training on a target data set (fine-tuning) or directly produce forecasts (zero-shot)~\citep{Pan.2009}.
Recent studies have shown that transfer learning can successfully enhance prediction performance or reduce computational and data requirements, compared to training a model from scratch on the target data. The related references can be divided into two approaches: pre-train on a similar source data set from the application perspective (specifically pre-trained model) to forecast a specific target data set, e.g.~\citet{Oreshkin.2021},~\citet{Wellens.2022}, or pre-train on a huge and diverse set of time series (foundation model~\citep{Bommasani.2022}) to supply a general foundation for forecasting any target data set, e.g.~\citet{Das.2024},~\citet{Goswami.2024}. 

Source-target similarity and source diversity are selection criteria for the source data sets of pre-trained models in time series forecasting, see for example~\citep{Oreshkin.2021} and~\citep{Rasul.2024}. Though, existing contributions explain little on how the choice of the source data set impacts transfer learning success, i.e., better zero-shot or fine-tuned forecasting performance than from scratch, on diverse target data sets. Especially if source and target data sets stem from different domains, transfer learning success might look like a lucky strike, leaving it unclear whether it can be transmitted to other use cases.
Few existing studies compare the similarity of different source data sets to the targets. ~\citet{He.2019}, who show that source data sets that favor the target in Dynamic Time Warping (DTW) provide better fine-tuned accuracy. However, to the best of our knowledge, no published study yet examines the relation between feature-based similarity measures~\citep{Alcock.1999} regarding time series characteristics e.g. trend or auto-regression of series in the source and target data sets, and the transfer learning forecasting performance.
While foundation models are built on massive diverse source data sets, there is no standardized definition of data set diversity and papers comparing sources of different diversity for transfer learning are lacking.

In contrast, most studies which analyze transfer learning results, focus on model architectures, hyper-parameters and training procedures and this either for the zero-shot or fine-tuned case (e.g.~\citet{Oreshkin.2021},~\citet{Kambale.2023},~\citet{Feng.2024}). A possible reason for this lack might be the seemingly limited choice of potential similar source data sets in applications and the method of selecting as many source data sets as possible to built a foundation data set. However, it is important to fill these gaps to provide more justification to pre-trained models and formulate indications for the success of transfer learning in new use cases. This can then be used to build targeted public time series databases. 

\begin{figure}[ht]
  \centering
  \includegraphics[width=500pt]{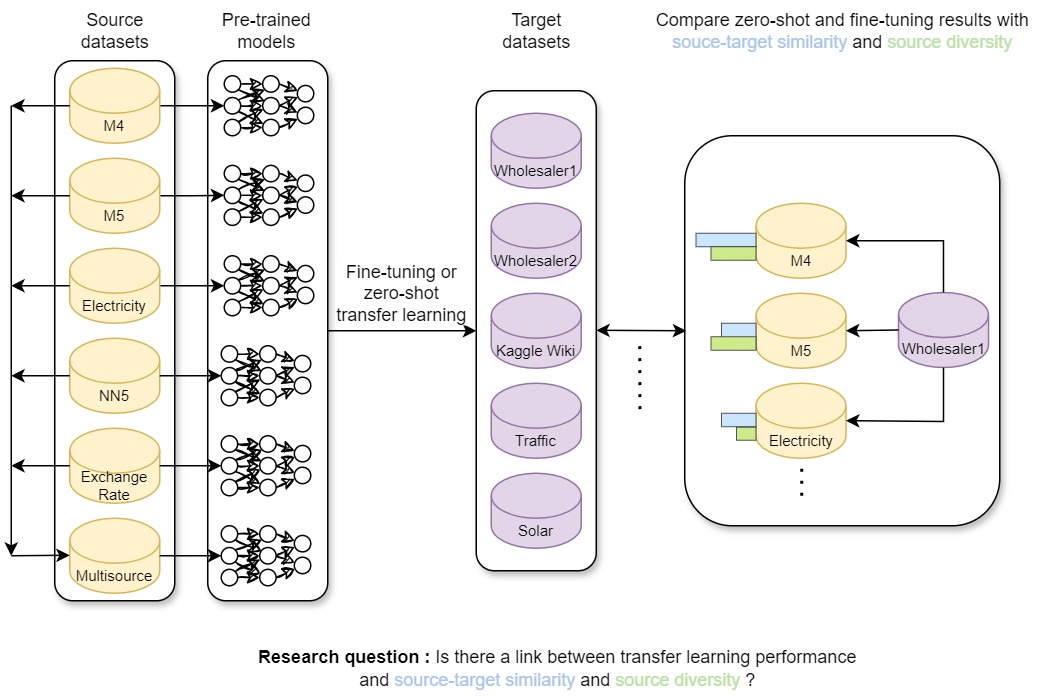}
  \caption{Summary of our approach and research question.}
  \label{graphical_abstract}
\end{figure}

Figure~\ref{graphical_abstract} summarizes our research question and approach. First, we pre-train a probabilistic recurrent neural network  DeepAR~\citep{Salinas.2020} on five publicly available source data sets of different domains and sizes. Additionally, we pre-train it on a concatenation of all these sets, comprising around 7.5 million data points, to create a small foundation model with respect to the training set, which we call the \texttt{Multisource} model.
We apply the pre-trained models in a zero-shot and fine-tuned manner to forecast five target data sets, of which three are public and two are non-public sales data, originating from different wholesalers. We measure the resulting performance in terms of forecasting accuracy, bias, and uncertainty estimation and benchmark the performance with \texttt{ETS} \citep{Rforecast}. To identify data set characteristics, we analyze the data at hand by calculating two different feature sets. Based on these as well as using a shape-based approach, we quantify similarities between the sources and targets. The features also serve as basis for the computation of source data diversity, which we define as the variance of time series features of the series within a data set. Additionally, we display data set diversity as time series features in the latent principal component analysis (PCA) space. Then, we map the forecasting performance on the target data to the similarity and diversity results using linear regression. We find that higher source diversity in catch22 features leads to more accurate point forecasts and uncertainty estimates on the target data set. Source-target similarity in tsfresh features leads to bias reduction, while the source diversity in tsfresh features increases it. These relationships are much stronger for the zero-shot than the fine-tuned forecasting results. We did not identify any consistent relation between forecasting performance and the regarded shape-based similarity metrics. 

Our contribution is two-fold:
\begin{itemize}
    \item we are the first to provide a structured investigation on how the source-target similarity and source diversity relates to zero-shot and fine-tuned transfer learning forecasting accuracy, bias and uncertainty estimation for data sets of different domains. 
    \item we provide the code used in this study open source (upon acceptance), containing methods for transfer learning, feature- and shape-based measures for calculating data set characteristics, relations between source-target similarity and source diversity and transfer learning performance as well as the pre-trained models to enable reproduction of the results and encourage further research.
\end{itemize}

The rest of the paper is structured as follows. First, we look at related work from the two lines of research, specifically pre-trained and foundation models. We highlight successful transfer learning experiments and the respective motivations behind the source and target data set selection. In Section \ref{sec: exp1}, we describe the experimental design of our transfer learning study, including models, data sets and metrics. After that, we show the results in terms of accuracy, bias, and uncertainty estimation in Section \ref{sec:exp1_results}. Then we present the similarity and diversity for the data at hand in Section \ref{sec: methods2}, and evaluate their relationship to forecasting performance in Section \ref{sec: relations_results}. In the conclusion we summarize and show directions for further research (Section \ref{sec:conclusion}).

\section{Related work}  
\label{sec:related_work}
While the recent success of conversational AI models is the result of research on foundation models for natural language processing, the discussion about transfer learning and pre-trained models frequently omits time series analysis. Only in recent years, the interest in transfer learning has conveyed to time series  -- for a review see~\citet{Weber.2021}. In the following, we summarize related work for time series analysis and forecasting, showing the potential of pre-trained models and the motivation behind the selected source and target data sets. While one stream of literature focuses on finding a similar source data set for a given target data set, the other stream aims to provide a general foundation model. 

\subsection{Specifically pre-trained models}
Pre-trained models have particular justification in forecasting applications, when the training data set is small, for example when new products or technical plants are introduced: In food retails, pre-trained models on a similar product can improve accuracy and enhance uncertainty estimates compared to the scratch model of a new product~\citep{Karb.2020}. The forecast of the 24 hours energy production of a new photovoltaic plant with a fine-tuned LSTM pre-trained on a nearby plant's data, is superior to training from scratch or zero-shot forecasting~\citep{Miraftabzadeh.2023}. Furthermore, pre-trained models can also be competitive in forecasting accuracy of normal sized data sets. \citet{Oreshkin.2021} demonstrate that a pre-trained N-BEATS~\citep{Oreshkin.2020} on the M4 data or FRED data is competitive against statistical methods for forecasting various data sets without even fine-tuning. They train one source model for each frequency of the M4 data and mostly choose the source model according to the target frequency.

Apart from the reduced data requirements, pre-trained models can also save training time:~\citet{Wellens.2022} show that the M5 competition winning LightGBM architecture pre-trained on more aggregated hierarchies and only fine-tuned at less aggregated ones of the M5 data set, can save 25\% of training time while keeping similar accuracy levels, compared to training from scratch.

Given a specific target data set, modelers aiming for pre-training seek appropriate, similar source data sets. Similarity is thereby often measured by apparently common features, like the application domain e.g.~\citep{Karb.2020}, the data set e.g.~\citep{Wellens.2022}, or the frequency e.g.~\citep{Oreshkin.2021}. Less papers examine similarities in the shape or the statistical characteristics of the time series within the source and the target data sets. In time series classification, pre-trained models on data sets of similarly shaped time series within the UCR archive~\citep{UCRArchive}, measured by the DTW, help target classification accuracy~\citep{IsmailFawaz.2018}. For small target data sets, transferring knowledge of data within the same domain help the classification task, even if the pre-training task is different~\citep{Ismailfawaz.2024}.~\citet{He.2019} show an analogical result for financial forecasting: source data sets consisting of two similar data sets, according to the DTW distances to the target data, provide more accurate fine-tuned forecasts than those combined from dissimilar sources. Furthermore, sources from the same industry have a positive effect on the accuracy, as well as using two instead of only one source data sets.

Pre-training with similar source data is also possible with more than one data set. This can be done by either combining the data to one big source and pre-training the model on it (we call it \texttt{Multisource} model) or by pre-training one model per data set and combining the models to an ensemble (we call it \texttt{Ensemble} model).~\citet{Gu.2021} for example, weight the pre-trained models per source according to the target similarity measured by the Maximum Mean Discrepancy and inter source data dependencies to gain a weighted ensemble.

While pre-trained models can increase target forecasting accuracy and uncertainty estimation, a small source data set can also lead to over-fitting, which decreases zero-shot performance on the target~\citep{Woo.2023}. In addition to using a source model of different complexity, pre-training on another similar data set of different size~\citep{Nakkiran.2021} or fine-tuning foundation models offer a remedy. 

\subsection{Foundation models}
Foundation models are trained on a huge and diverse set of data with the aim to build a foundation for any downstream task. Accordingly, selecting a specific source data set is unnecessary. However, until recently the success of time series foundation models was limited, which might have been due to the lack of a massive structured public time series repository, like those existing for text and images, to train a large source model~\citep{Hahn.2023}.

The first foundation model for time series forecasting, which gained a lot of attention, was the proprietary TimeGPT-1~\citep{Garza.2023}, a transformer based deep learning architecture, pre-trained on a vast amount of time series data from diverse domains. TimeGPT-1 reveals superior performance on monthly and weekly data and good performance on daily and hourly data. Yet, the authors do not reveal their training and test data sets and stay reserved about the details of their architecture and training parameters.
More transparent studies have emerged on foundation models trained on the Monash Time Series Repository~\citep{Godahewa.2021}, which encompasses real-world time series data of diverse domains. The Lag-Llama transformer model of~\citet{Rasul.2024} e.g., which was pre-trained on it, beats all the statistical and machine learning benchmark models and another pre-trained transformer, when zero-shot forecasting the held-out Traffic~\citep{Lai.2018} data set. The authors show that the zero-shot performance of Lag-Llama on the held out data sets increases with the number of samples used for pre-training. Additionally, they visualize the diversity in time series characteristics of the source data sets by plotting the mean catch22~\citep{Lubba.2019} feature vector per data set in the PCA space. Data sets of the same domain are projected into similar regions, suggesting that source data sets consisting of different domains, are more diverse. They call for further research on how the source diversity relates to the forecasting performance, and we address this in our work.~\citet{Ekambaram.2024} also used the Monash repository to pre-train the first publicly available forecasting foundation model, which can be used for general time series forecasting. Specifically, it allows for multivariate forecasting, the integration of external variables and different frequencies. Their pre-trained zero and few-shot models show comparable or superior performance against state-of-the-art pre-trained large language models, pre-trained forecasting models and traditionally trained machine learning models on uni- and multi-variate data sets with and without exogenous variables which were not part of the pre-training.

Foundation models can also be built out of general and diverse synthetically generated time series. For example~\citet{Dooley.2024} generate 300,000 time series containing multiplicative trend, seasonal and error components to train a transformer model for zero-shot forecasting. The resulting model outperforms the benchmark models like naive, transformer-based scratch models as well as the pre-trained NBEATS model, in terms of the MSE on different forecast horizons and seven benchmark data sets. Using pre-trained models on artificial data can be better than searching for an existing source data set. This can also be the case if the pre-training task is different to the target task:~\citet{Rotem.2022} created 15 million time series on which they pre-trained a CNN for time series regression. After fine-tuning for classification on target data from the UCR archive, this source model works better than selecting a source data set from the UCR archive, when regarding only 10\% for training or fine-tuning on the target.

Overall, research suggests that the success of a foundation model depends source wise on the size and the diversity of the source data. Foundation models are specifically competitive when there is limited data in the target data set or training time. 

This study expands the current research by structurally examining the relationships between source diversity, source-target similarity, and forecasting performance on the target. It integrates models pre-trained on single source data sets, their ensemble, and a small foundation model trained on the combination of five data sets from diverse domains.

\section{Transfer learning experiment} 
\label{sec: exp1}
We first explore the forecasting performance of pre-trained models, where the source and target data stem from different domains. To that end, we select five diverse source and target data sets and divide them each into training and test set. We use the training set of each source to pre-train source models and additionally train the \texttt{Multisource} model on the concatenation of the sources' training sets. Note that this training data set is relatively small in comparison to those of state-of-the-art foundation models, since we aim to analyze the relation of the data set characteristics to the forecasting performance and limit the computational effort. Still, it can be seen as a small foundation model. Furthermore, we build an ensemble of the source models, the \texttt{Ensemble} model. We evaluate the pre-trained models with regard to their forecasting performance on the test sets of the target data sets for the zero-shot and fine-tuned case. For the former, we directly forecast with the pre-trained models and for the latter we first fine-tune the models on the target training sets. We benchmark the forecasting performance of the pre-trained models against the so-called \texttt{Scratch} model, which uses the training data of the target data sets to train from scratch, i.e. with a random start initialization.  This comparison provides information about the added value of transfer learning. Also, we include the ETS models as benchmark. The details of the experiment are given in the following.

\subsection{Source data}
As source data for pre-training, we select five \emph{publicly available} data sets from the Monash Time Series Forecasting Repository~\citep{Godahewa.2021} which are \emph{diverse regarding size and domain}: \texttt{M5}~\citep{Makridakis.2022}, \texttt{M4}~\citep{Makridakis.2020}, \texttt{Electricity}~\citep{Lai.2018}, \texttt{NN5}~\citep{Crone.2009} and \texttt{Exchange Rate}~\citep{Lai.2018}. We directly use the available weekly \texttt{NN5} and \texttt{M4} data sets and otherwise aggregate the time series to weekly frequency. To keep the scope of the experiment manageable, we disregard other frequencies and restrict to univariate time series. The chosen source data sets and their size are listed in Table~\ref{source1}. For each time series, we use 80\% of the time steps as training set to pre-train the source models. Overall, we use 31,300 time series, and in sum around 7.5 million observations for pre-training.

\begin{table}[ht]
\footnotesize
\begin{threeparttable}
\begin{tabular}[t]{L{0.17\textwidth} L{0.12\textwidth}L{0.17\textwidth}R{0.13\textwidth}R{0.14\textwidth}R{0.13\textwidth}} 
      \hline
        Data set &  Domain & Description & \# Time Series & \# Training Periods & Training Size\\ 
     \midrule
     \texttt{M5} & Sales & Walmart SKU sales & 30,490 & 219 & 6,677,310\\
     \texttt{M4} & Diverse\tnote{a} & Diverse series & 359 & 2,000 & 718,000\\
     \texttt{Electricity} & Energy & Electricity consumption & 321 & 102 & 32,742\\
     \texttt{NN5} & Banking & Cash withdrawals at ATMs & 111 & 90 & 9,990\\ 
     \texttt{Exchange Rate} & Banking & Exchange rates & 8 & 996 & 7,968\\  
     \hline
     \textbf{Sum} &  &  & 31,289 & 3,407 & 7,446,010\\
     \hline
\end{tabular}
\begin{tablenotes} \footnotesize
	\item[a] e.g. Micro, Macro, Industry, Finance
\end{tablenotes}
\end{threeparttable}
\caption{Source data sets sorted by training size.} \label{source1}
\end{table}

\subsection{Target data}
To test the transferability of knowledge stored in pre-trained models, we select three publicly available target data sets based on the criteria applied to select source data sets: \texttt{Kaggle Wiki }~\citep{Anava.2017}, \texttt{Traffic}~\citep{Lai.2018} and \texttt{Solar Energy}~\citep{Lai.2018}. In addition, we consider two SKU wholesales project data sets from two different wholesalers, \texttt{Wholesaler1} and \texttt{Wholesaler2}, a domain without well-known public data. Again, we select the data sets aggregated to weekly frequency, if available or aggregate by ourselves. Table~\ref{target1} features a short description per target data set. Note that the target domains differ from those of the source data to examine transfer learning, data set similarity and diversity more broadly. The energy domain appears in the sources and in the targets, but the underlying business processes differ. 

Again, we split the data sets 80:20 along the time axis to train the \texttt{Scratch} model or fine-tune the pre-trained model, and test the model. To keep the computational effort of the experiment manageable, we restrict the target test data sets for evaluation to a maximum of 1,000 randomly drawn time series each. For data sets containing fewer time series, we account for the entire data set. We evaluate the forecasts on rolling origins~\citep{Tashman.2000} of the target test sets. For the \texttt{Solar Energy} data set, 20\% of the time steps would be less than the forecast horizon. Since we need at least one forecast per time series for evaluation, we enlarge the test set here.

\begin{table}[ht]
\footnotesize
\begin{threeparttable}
\begin{tabular}[t]{L{2cm}L{3.5em}L{2.4cm}R{1.2cm}R{1.2cm}R{1.5cm} R{1.15cm} R{1.15cm} R{1.1cm}} 
 \hline
 Data set &  Domain & Description & Time Series & Training Periods & Training Size & Test Periods & Test Size & Rolling Origins \\ 
  \hline
 \texttt{Wholesaler1} & Sales & SKU sales & 113,000 & 130 & 14,690,000 & 32 & 32,000 & 18 \\ 
 \texttt{Wholesaler2} & Sales & SKU sales & 55,000  & 124 & 7,820,000 & 31 & 31,000 & 17 \\ 
 \texttt{Kaggle Wiki }& Web & Web traffic of Wikipedia pages & 145,000 & 91 & 13,195,000 & 23 & 23,000 & 9\\ 
 \texttt{Traffic} & Transport & Occupancy rate of car lanes & 862 & 68 & 58,616 & 17 & 14,654 & 3\\ 
 \texttt{Solar Energy} & Energy & Solar power production of a state & 137 & 35 & 4,795 & 15 & 2,055 & 1\\ \hline
 Sum & & & 313,999 & 448 & 35,768,411 & 118 & 102,709 & 48 \\
 \hline
\end{tabular}
\end{threeparttable}
\caption{Target data sets sorted by test size.} \label{target1}
\end{table}

\subsection{Transfer learning and scratch models}
We apply the neural network architecture DeepAR~\citep{Salinas.2020} as the source and \texttt{Scratch} model architecture. DeepAR is well known in the forecasting community and can be trained in a global manner, i.e. over multiple time series to extract joint knowledge. Further, it is able to output probabilistic forecasts, i.e. give an estimation about the forecast uncertainty.

For robustness, we use an ensemble composed of five DeepAR models. Pre-trained and \texttt{Scratch} models share the hyper-parameters, listed in Table \ref{hyperparams}. The models encompass as many output nodes as the forecast horizon for direct forecasting. To save computational resources, we waive hyper-parameter optimization and take the default values of the implementations in the \textit{gluonts} package~\citep{gluonts_jmlr}. One exception is the output distribution of the DeepAR, where we select the negative binomial as proposed in~\citep{Salinas.2020}. The DeepAR model comes with a default item-wise mean scaling, which helps in adapting to the target data sets' scales. Hence, we do not scale the data during pre-processing.

Scratch training, pre-training and fine-tuning involve all layers and are conducted with the \textit{gluonts mx.trainer} with an early stopping adaptation, including a maximum of 800 epochs.

\begin{table}[ht]
\centering
\begin{tabular}{lrr } 
 \hline
  & DeepAR & ETS \\ 
  \hline
 Ensemble members & 5 & - \\ 
 Context length & 25 & 25\\ 
 Forecast horizon & 15 & 15 \\ 
 Output distribution & negative binomial  & - \\ 
 Number samples & 500 & 100\\ 
 Max epochs & 800 & - \\ 
 Number layers & 2 & - \\ 
 \hline

\end{tabular}
\caption{Hyperparameters for the source, scratch, and benchmark model.}
\label{hyperparams}
\end{table}

\subsection{ETS benchmark model}
As representative local statistical benchmark, we use the \texttt{ETS} model via the \textit{gluonts} R wrapper, \textit{r\_forecast}, which uses the R \textit{forecast} package~\citep{Rforecast}. There, the default parameters are set, which corresponds to an automatically selected error, trend and season type, an estimation of the model parameters and trying both, damped and non-damped trend. Depending on the model type, prediction intervals are gained according to closed form forecast variance formulas if available for the respective ETS model or simulated sample paths otherwise ~\citep{Hyndman.2021}.

\subsection{Metrics}
\label{metrics}
For the evaluation of forecast performance, we take the accuracy, bias, and uncertainty estimation into account.
For any target data set, we denote the number of time series with $m$, and the series themselves with  $y_i, i = 1,...,m$. The test set contains rolling origins $r = 1,...,R$, which start after the past time steps $t=1,...,T_r$ and encompass the test time steps $t=T_r+1,...,T_r+h$ during the forecast horizon $h$. The accuracy and bias measure refers to the median forecast $\hat{y}_{i,t}$ of a time series $i$ at time $t$. The uncertainty estimation error refers to the forecast quantiles $\hat{q}_{i,j,t}$, where $j$ is the quantile percentage of time series $i$ at time $t$.

For the accuracy we consider the Average Root Mean Squared Scaled Error (AvgRMSSE) given in Eq.~\ref{AvgRMSSE} which is based on the RMSSE~\citep{Makridakis.2022}. It is the average over rolling origins and time series of the forecast RMSSE in Eq.~\ref{RMSE_forecast} relative to the naive forecast RMSE in the training data Eq.~\ref{RMSE_naive_past}. 

\begin{align}
AvgRMSSE &= \frac{1}{R} \sum_{r=1}^{R} \left( \frac{1}{m} \sum_{i=1}^{m}(RMSE_{i}^{r}/RMSE_{i,naive past}^{r}) \right) \label{AvgRMSSE}\\
RMSE_{i}^{r} &= \sqrt{ \frac{1}{h} \sum_{t=T_r+1}^{T_r+h} (y_{i,t} - \hat{y}_{i,t})^2} \label{RMSE_forecast}\\
RMSE_{i,naive past}^{r} &= \sqrt{\frac{1}{T_r-1} \sum_{t=2}^{T_r} (y_{i,t} - y_{i,t-1})^2} \label{RMSE_naive_past}
\end{align}

The bias is measured by the Mean Error (ME) in Eq.~\ref{me}, which averages the mean error of the forecast in Eq.~\ref{me_forecast} over rolling origins and time series.

\begin{align}
ME &=  \frac{1}{R} \sum_{r=1}^{R} \left( \frac{1}{m} \sum_{i=1}^{m}(ME_{i}^{r})\right) \label{me}\\
ME_{i}^{r} &= \frac{1}{h} \sum_{t=T_r+1}^{T_r+h} (y_{i,t} - \hat{y}_{i,t}) \label{me_forecast}
\end{align}

We assess the uncertainty estimation of the sample forecasts by a modified Mean Scaled Interval Score (MSIS)~\citep{Makridakis.2020}, given in Eq.~\ref{msis}. It averages the MSIS with $\alpha = 0.05$ of each forecast in Eq.~\ref{MSIS_forecast} over time series and rolling origins. For consistency in scaling, we use the RMSE of the naive forecast in the past (Eq.~\ref{RMSE_naive_past}) instead of its MAE, as proposed by the authors, in the denominator. 

\begin{align}
MSIS &= \frac{1}{R} \sum_{r=1}^{R} \left( \frac{1}{m} \sum_{i=1}^{m}(MSIS_{i}^{r})\right) \label{msis}\\
MSIS_{i}^{r} &=  \frac{\frac{1}{h} \sum_{t=T_r+1}^{T_r+h} ( \hat{q}_{i,97.5,t} - \hat{q}_{i,2.5,t}) + \frac{2}{0.05} (\hat{q}_{i,2.5,t} - y_{i,t}) \mathds{1}_{(y_{i,t} < \hat{q}_{i,2.5,t})} + \frac{2}{0.05} (y_{i,t} - \hat{q}_{i,97.5,t}) \mathds{1}_{(y_{i,t} > \hat{q}_{i,97.5,t})}}{RMSE_{i,naive past}^{r}} \label{MSIS_forecast}
\end{align}

We report fine-tuning times of pre-trained models in relation to scratch training times on a Linux virtual machine with 32 CPUs and 77 GB RAM in the Appendix Table~\ref{times}. For that, we use the python package \textit{codecarbon}~\citep{codecarbon}.

\section{Result analysis} 
\label{sec:exp1_results}
In the following, we provide the accuracy, bias, and uncertainty estimation performance of the pre-trained, \texttt{Scratch} and benchmark model.

\begin{table}[ht]
    \centering
    \hspace*{-1cm}
    \begin{tabular}{l|rrrrr}
    \hline
        Source / Target & \texttt{Solar} & \texttt{Traffic} & \texttt{Kaggle Wiki } & \texttt{Wholesaler2} & \texttt{Wholesaler1} \\ \hline
        \texttt{Multisource} & 0.75/0.78 & 3.40/2.13 & \underline{\textbf{0.99}}/\underline{\textbf{1.03}} & 0.92/\underline{1.00} & \underline{0.81}/\underline{0.80} \\ 
        \texttt{Ensemble} & 0.86/\underline{\textbf{0.75}} & 2.10/1.96 & 1.32/1.06 & 1.08/1.06 & 4.20/1.36 \\ 
        \texttt{M5} & 1.45/0.77 & 4.46/\underline{1.60} & 1.42/1.06 & 1.09/1.09 & 2.93/1.21 \\ 
        \texttt{M4} & \underline{\textbf{0.71}}/0.76 & \underline{1.97}/2.52 & 1.23/\underline{\textbf{1.03}} & \underline{\textbf{0.77}}/1.01 & 3.49/1.15 \\ 
        \texttt{Electricity} & 1.02/0.85 & 7.72/2.62 & 1.66/1.04 & 1.31/1.05 & 5.93/0.95 \\ 
        \texttt{NN}5 & 0.77/0.77 & 3.20/2.34 & 1.29/1.16 & 1.25/1.16 & 4.36/1.26 \\ 
        \texttt{Exchange Rate} & \underline{\textbf{0.71}}/0.79 & \underline{1.97}/1.72 & 1.38/1.09 & 1.08/1.03 & 4.53/2.32 \\ \hline
        \texttt{Scratch} & 0.85 & 3.04 & 1.05 & 1.19 & 1.74 \\ \hline
        \texttt{ETS} & 0.96 & \textbf{1.08} & 1.11 & \textbf{0.79} & \textbf{0.60} \\ \hline

    \end{tabular}
    \caption{Zero-shot/fine-tuned transfer learning, scratch and benchmark accuracy performance in AvgRMSSE (see Eq.~\ref{AvgRMSSE}). Smaller values are better. The best results for each target data set and mode are printed in bold and best results within the pre-trained models are underlined.}
\label{acc_results}
\end{table}

\subsection{Accuracy}
Table~\ref{acc_results} shows the accuracy measured in terms of the Average RMSSE (see Eq.~\ref{AvgRMSSE}) of the pre-trained, \texttt{Scratch} and benchmark models for the different source and target combinations. There is always at least one pre-trained model that beats the \texttt{Scratch} model in the zero-shot case, i.e. without having ever seen the target data and which has been trained on data of another domain. In 40\% of the time zero-shot pre-trained models beat the \texttt{Scratch} model. If a pre-trained model provides a zero-shot performance that is superior to the \texttt{Scratch} model, only in 35\% it gets even better with fine-tuning. The \texttt{ETS} model is the best fit for \texttt{Traffic} and \texttt{Wholesaler1} and cannot be beaten even after fine-tuning.

The fine-tuned pre-trained models beat the \texttt{Scratch} model in 89\% of the cases, showing the potential of pre-trained models. Note that the shared hyperparameters of pre-trained and \texttt{Scratch} models are default, making their practical usage realistic. This is in contrast to other papers comparing huge architectures used for pre-training also for \texttt{Scratch} model training on a single data set. Within the pre-trained models, the \texttt{Multisource} and \texttt{M4} source models achieve the best transfer learning accuracy. Fine-tuning enhances the accuracy of the pre-trained models in 74\% of the cases. This particularly applies to the \texttt{Electricity} source data set, where accuracy improves from the worst in zero-shot forecasting to intermediate after fine-tuning. In contrast, the \texttt{Multisource} and the \texttt{M4} source models tend to get worse with fine-tuning. A possible reason might be that the model parameters converge against those of the \texttt{Scratch} model during fine-tuning. If this is worse than the zero-shot performance, the fine-tuned performance drops. While the \texttt{M5} data is a huge part of \texttt{Multisource} (89\%), their respective source models provide distinct accuracy measures on the targets. The \texttt{Multisource} model displays a superior performance than the \texttt{Ensemble}, although both were trained on the same data. In contrast to the other target data sets, pre-training on the \texttt{M5} source provides better zero-shot performance on the wholesales data sets, which might be due to similar time series characteristics in retail and wholesales. This is also in line with the literature in Section \ref{sec:related_work}.
 
\begin{figure}[ht]
  \centering
  \includegraphics[width=500pt]{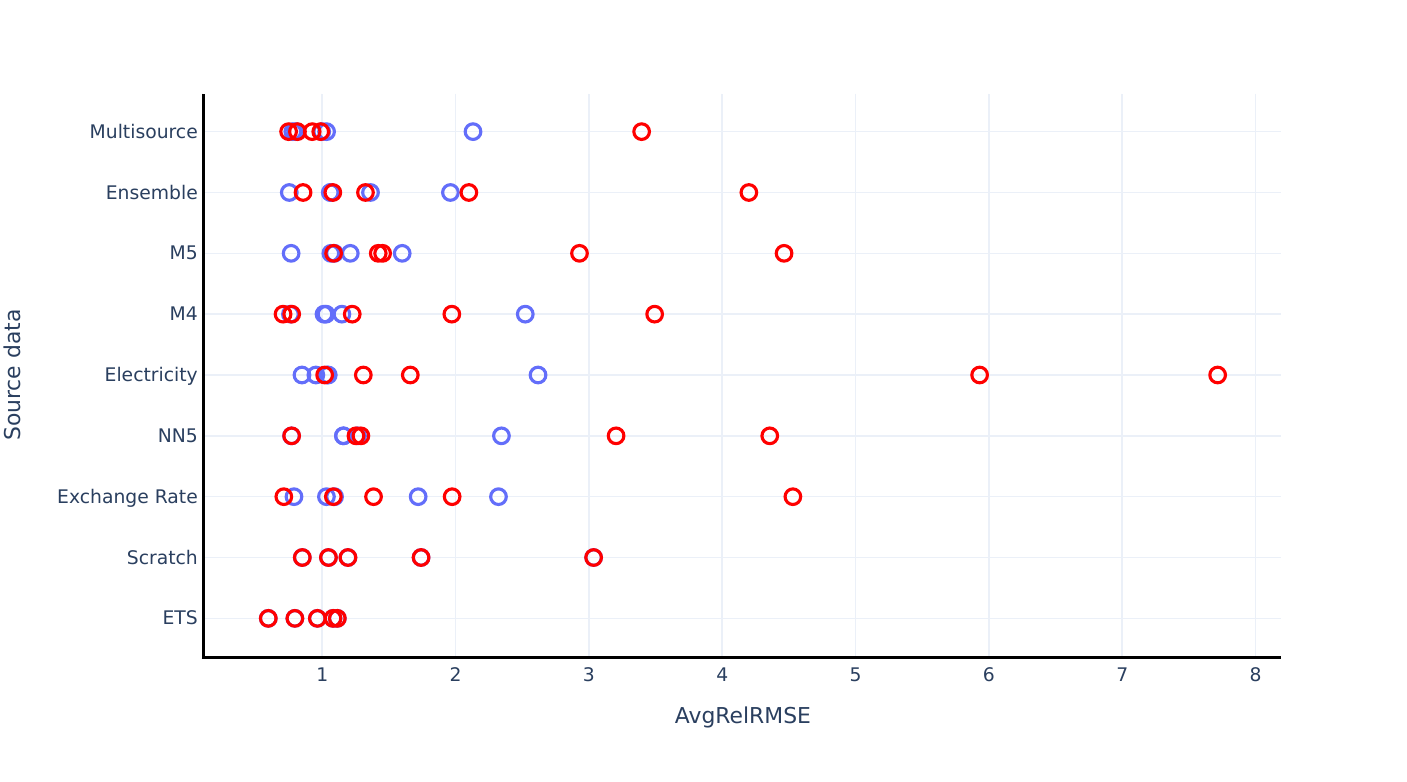}
  \caption{Zero-shot/scratch (red circles) and fine-tuned (blue circles) accuracy in AvgRMSSE (see Eq.~\ref{AvgRMSSE}) per source model across the five target data sets. View in color.}
  \label{acc_over_targets}
\end{figure}

Figure \ref{acc_over_targets} compares each model's zero-shot and fine-tuned accuracy given one source data set across the different target data sets. This view demonstrates the overall effect of fine-tuning leading to enhanced accuracy. The model pre-trained on the \texttt{Multisource} shows similarly good performance in both cases and on four out of five target data sets.  

To summarize: comparing the pre-trained models, those pre-trained on the \texttt{Multisource} and the \texttt{M4} data sets provide superior accuracy performance and fine-tuning enhances the performance in general.

\subsection{Bias}
\begin{table}[ht]
\centering
\hspace*{-1cm}
    \begin{tabular}{l|rrrrr}
    \hline
        Source / Target & \texttt{Solar} & \texttt{Traffic} & \texttt{Kaggle Wiki } & \texttt{Wholesaler2} & \texttt{Wholesaler1} \\ \hline
        \texttt{Multisource} & -1.08/-0.08 & +1.12/+0.48 & -0.55/+1.14 & +0.62/+1.96 & +1.30/3.04 \\ 
        \texttt{Ensemble} & -0.65/-0.26 & +0.73/+0.34 & -0.24/+0.41 & +0.61/+1.65 & +1.12/+2.26 \\ 
        \texttt{M5} & -2.53/\underline{\textbf{+0.04}} & +1.38/+0.44 & -2.52/+0.93 & \underline{\textbf{+0.01}}/+2.04 & +0.75/+3.01 \\
        \texttt{M4} & -0.28/-0.15 & +0.59/+0.31 & +1.10/\underline{\textbf{-0.08}} & +1.51/+1.98 & +1.80/+3.01 \\ 
        \texttt{Electricity} & -1.40/-1.19 & +4.82/+0.38 & -0.25/+0.93 & +2.10/+1.94 & +2.18/+2.77 \\ 
        \texttt{NN5} & -1.57/+0.31 & \underline{\textbf{-0.01}}/\underline{\textbf{+0.22}} & -0.61/-0.29 & -0.19/+1.65 & \underline{\textbf{+0.63}}/+2.25 \\ 
        \texttt{Exchange Rate} & \underline{\textbf{+0.14}}/-0.36 & +0.74/+0.34 & \underline{+0.10}/+0.59 & +0.19/\underline{\textbf{-0.15}} & +0.87/\underline{\textbf{+0.68}} \\ \hline
        \texttt{Scratch} & +0.94 & +0.25 & \textbf{-0.09} & +2.06 & +2.86 \\ \hline
        \texttt{ETS} & -2.27 & +0.26 & -1.75 & -0.24 & +0.89 \\ \hline
    \end{tabular}
\caption{Zero-shot/fine-tuned bias performance in scaled Mean Error. A good performance in this measure corresponds to a bias close to zero. Negative values stand for underestimating forecasts and positive for overestimating. The best source data set for each target data set and mode is printed in bold and best results within the pre-trained models are underlined.}
\label{bias_results}
\end{table}

Table \ref{bias_results} reports the scaled bias in terms of the Mean Error (see Eq.~\ref{me}) for the zero-shot and fine-tuned case. To compare the zero-shot and fine-tuned bias, we scale the bias over the results of both modes per target data set, by dividing by the bias standard deviation. For both modes, with one exception, there is always a pre-trained model which reveals a better bias than the \texttt{Scratch} and the \texttt{ETS} model. 
Overall, fine-tuning reduces the forecasting bias in 51\% of the time. Within the pre-trained models, the one pre-trained on the \texttt{Exchange Rate} provides the lowest bias.

To summarize: depending on the source data set, pre-trained models reduce bias compared to the \texttt{Scratch} and the benchmark model.

\subsection{Uncertainty estimation}
\begin{table}[ht]
\centering
\hspace*{-1cm}
    \begin{tabular}{l|rrrrr}
    \hline
        Source / Target & \texttt{Solar} & \texttt{Traffic} & \texttt{Kaggle Wiki } & \texttt{Wholesaler2} & \texttt{Wholesaler1} \\ \hline
        \texttt{Multisource} & 4.67/5.28 & 63.47/37.90 & \underline{7.32}/6.72 & \underline{\textbf{6.67}}/7.37 & \underline{5.88}/\underline{6.00} \\ 
        \texttt{Ensemble} & 10.52/4.70 & 117.59/38.22 & 10.11/6.67 & 7.15/\underline{\textbf{7.27}} & 19.68/7.66 \\ 
        \texttt{M5} & 16.2/6.14 & 65.11/41.96 & 10.73/6.70 & 7.35/8.18 & 17.80/9.47 \\ 
        \texttt{M4} & 4.42/5.85 & 60.75/38.90 & 8.09/\underline{\textbf{6.25}} & 8.06/7.62 & 14.12/8.46 \\ 
        \texttt{Electricity} & 10.97/7.66 & 253.25/37.01 & 17.13/6.49 & 9.82/7.79 & 42.05/7.58 \\ 
        \texttt{NN5} & \underline{4.18}/\underline{4.39} & 33.66/37.12 & 13.18/7.51 & 9.28/8.03 & 51.87/6.86 \\ 
        \texttt{Exchange Rate} & 8.11/4.84 & \underline{32.37}/\underline{34.38} & 29.35/7.29 & 9.75/7.94 & 104.24/26.49 \\ \hline
        \texttt{Scratch} & \textbf{4.01} & 38.11 & \textbf{6.67} & 9.45 & 15.48 \\ \hline
        \texttt{ETS} & 5.38 & \textbf{7.40} & 7.71 & 7.58 & \textbf{5.11} \\ \hline
    \end{tabular}
\caption{Zero-shot/fine-tuned uncertainty estimation performance in MSIS (see Eq.~\ref{msis}). Smaller values are better. The best source data set for each target data set and mode is printed in bold and best results within the pre-trained models are underlined.}
\label{uncertainty_results}
\end{table}

Table \ref{uncertainty_results} shows the zero-shot and fine-tuned uncertainty estimation performance in terms of the MSIS (see Eq. \ref{msis}). For the \texttt{Solar}, \texttt{Traffic} and \texttt{Wholesaler1} data sets, either the \texttt{Scratch} or the \texttt{ETS} provides better results than the pre-trained models.

The results resemble the accuracy results for point forecasts in Table \ref{acc_results}: within the pre-trained models, the \texttt{Multisource} data set represents the best source data set for transfer learning. Similar to the accuracy, the uncertainty estimation usually improves with fine-tuning, in 74\% of the cases. While the zero-shot performance of the \texttt{Electricity} source model is inferior to other pre-trained models, it catches up with fine-tuning. 
To summarize: comparing the pre-trained models, those pre-trained on the \texttt{Multisource} data set provides superior uncertainty estimation performance and fine-tuning enhances the performance in general.

\section{Interrelations between source and target data sets} 
\label{sec: methods2}
The overall goal of this paper is to find relations between source-target similarity, source diversity, and transfer learning performance in accuracy, bias and uncertainty estimation. Having evaluated the forecasting performances for zero-shot and fine-tuned pre-trained models in Section \ref{sec:exp1_results}, we turn to measuring data set similarity and diversity. This then allows us to map these characteristics of the source and target data sets to the performance of transfer learning in the next section.

We compute two different time series feature sets as well as averaged time series for all source and target data sets. Based on these, we calculate two feature-based and one shape-based similarity measure, and two feature-based diversity measures.
As feature-based similarity measure, we use the median Euclidean feature distances between all sources and targets. We express the feature-based diversity per data set by the feature variance and visualize this in the latent principal component space. We also include DTW distances of the data set representatives. In the following, we describe the time series features, similarity and diversity measures in more detail. 

We restrict the number of time series considered per data set for these calculations by randomly drawing 1,000 of their ids. If a data set contains less than this number of time series, all are taken into account. We limit the number of time series to save computational resources while still producing representative results.

\subsection{Time series features}
\label{sec:TS Features}
We extract ten common time series features to characterize the data via the \textit{tsfresh} package~\citep{Christ.2018}: "absolute energy", "intermittency", "mean", "median", "kurtosis", "skewness", "standard deviation", "aggregated autocorrelation", "lumpyness" and "aggregated linear trend". This selection is parsimonious and covers both scale and shape characteristics. Further details on the extracted features can be found in the Appendix in Table~\ref{tsfresh_feat}.
To not limit our analysis to a single hand-selected feature set, and since catch22 features~\citep{Lubba.2019} are known to be a helpful input for time series classification, we further extract catch22 features via the pycatch22 toolkit \footnote{https://github.com/DynamicsAndNeuralSystems/pycatch22}. These include e.g. key figures on the distribution, temporal statistics and linear and non-linear autocorrelation. We use all 22 catch22 features plus the "mean" and "standard deviation".

\subsection{Feature-based data set similarity}
To assess the similarity of source and target data sets, we z-transform the extracted features over all source data sets and separately over all target data sets, so that they have zero mean and unit variance per feature. Let $f_{i,s_n}$ be the standardized feature vector of the $i$-th time series of the source data set $s_n$ and $f_{j, t_m}$ the standardized feature vector of time series j of the target data set $t_m$, either for tsfresh or catch22. 
Based on that, we calculate the Euclidean distance between each feature $p$ of each source and target time series pair, $dist(f_{i,s_n}[p],f_{j, t_m}[p])$.
Then, we compute the median over all time series pairs' distances per feature $p$: $dist_p(s_n, t_m) = median(dist(f_{i,s_n}[p],f_{j, t_m}[p]))$. Finally, we compute the median over the feature distances for every source-target data set pair: $dist(s_n, t_m) = median(dist_p(s_n, t_m))$.
The resulting distance matrix $dist(s_n, t_m)$ provides the median feature distance between each source and target data set. We compute these distance matrices for both the tsfresh and the catch22 features.

\begin{table}[ht]
    \centering
    \begin{tabular}{l|rrrrr}
    \hline
        Source/Target & \texttt{Solar} & \texttt{Traffic} & \texttt{Kaggle Wiki} & \texttt{Wholesaler2} & \texttt{Wholesaler1} \\ \hline
        \texttt{Multisource} & 1.36 & 0.24 & 0.22 & 0.46 & 0.30 \\ 
        \texttt{M5} & 1.35 & 0.24 & 0.22 & 0.45 & 0.29 \\
        \texttt{M4} & 1.50 & 0.26 & 0.25 & 0.59 & 0.41  \\
        \texttt{Electricity} & \textbf{1.22} & 0.30 & 0.29 & 0.50 & 0.31 \\ 
        \texttt{NN5} & 1.64 & 0.21 & 0.20& \textbf{0.42} & \textbf{0.25} \\
        \texttt{Exchange Rate} & 1.23 & \textbf{0.19} & \textbf{0.18} & 0.46 & \textbf{0.25} \\ \hline
    \end{tabular}
    \caption{Median tsfresh feature distances of source target pairs. The distance values of the least distant i.e. the most similar source-target pairs are printed in bold.}
    \label{tsfresh_distances}
\end{table}

\begin{table}[ht]
    \centering
    \begin{tabular}{l|rrrrr}
    \hline
        Source/Target & \texttt{Solar} & \texttt{Traffic} & \texttt{Kaggle Wiki} & \texttt{Wholesaler2} & \texttt{Wholesaler1} \\ \hline
        \texttt{Multisource} & 0.82 & 0.73 & 0.65 & 0.63 & 0.67 \\ 
        \texttt{M5} & 0.82 & 0.71 & \textbf{0.64} & \textbf{0.61} & \textbf{0.65} \\ 
        \texttt{M4} & 1.22 & 1.03 & 1.07 & 1.25 & 1.1 \\ 
        \texttt{Electricity} & \textbf{0.59} & \textbf{0.61} & 0.65 & 0.97 & 0.91 \\ 
        \texttt{NN5} & 0.75 & 0.71 & 0.74 & 0.73 & 0.69 \\ 
        \texttt{Exchange Rate} & 1.29 & 1.09 & 1.05 & 1.22 & 1.17 \\ \hline
    \end{tabular}
    \caption{Median catch22 feature distances of source target pairs. The distance values of the least distant i.e. the most similar source-target pairs are printed in bold.}
    \label{catch22_distances}
\end{table}

Table \ref{tsfresh_distances} lists the median Euclidean feature distances between the source and target data sets for the tsfresh features, and Table \ref{catch22_distances} lists those for catch22. Smaller values indicate higher similarity. In terms of tsfresh feature distances, \texttt{NN5} or \texttt{Exchange Rate} is the closest to all target data sets and \texttt{Electricity} or \texttt{M4} is the most distant except \texttt{Solar Energy}, where it is the other way round. In terms of catch22 features, \texttt{Electricity} is the closest source data set to \texttt{Solar} and \texttt{Traffic} and otherwise it is the \texttt{M5}. This highlights that the feature sets capture fundamentally different time series characteristics. Hence, we keep them both for further analysis.

\subsection{Shape-based data set similarity}
To also account for shape-based similarity measures, we compute a representative time series of each source and target data set based on the DTW Bary Center Averaging (DBA) method~\citep{Petitjean.2011}. In that, we use the \textit{tslearn} package~\citep{Tavenard.2020}, which was implemented based on~\citet{Schultz.2018}. The time series are first scaled to zero mean and unit variance per data set. 

\begin{table}[ht]
    \centering
    \begin{tabular}{l|rrrrr}
    \hline
        Source/Target & \texttt{Solar} & \texttt{Traffic} & \texttt{Kaggle Wiki} & \texttt{Wholesaler2} & \texttt{Wholesaler1} \\ \hline
         \texttt{Multisource} & \textbf{0.69} & 2.48 & \textbf{0.19} & 3.31 & 3.29 \\ 
        \texttt{M5} & 5.35 & 4.23 & 6.03 & 4.03 & 3.78 \\ 
        \texttt{M4} & 27.01 & 26.13 & 28.79 & 9.51 & 8.96 \\ 
        \texttt{Electricity} & 4.02 & \textbf{2.20} & 4.55 & 3.31 & \textbf{3.09} \\ 
        \texttt{NN5} & 3.52 & 3.30 & 3.82 & 3.44 & 3.36 \\ 
        \texttt{Exchange Rate} & 7.84 & 7.37 & 8.62 & \textbf{3.07} & 3.80 \\ \hline
    \end{tabular}
    \caption{DTW distances of source-target pairs. The distance values of the least distant i.e. the most similar source-target pairs are printed in bold.}
    \label{dtw_distances}
\end{table}

The distances between the data sets are then assessed by the DTW~\citep{Petitjean.2011} distances of the data set representatives, implemented by DTAIDistance~\citep{Meert.2020}. They are provided in Table \ref{dtw_distances}. Here, either the \texttt{Electricity} or the \texttt{Multisource} are the most similar source for all target data sets, except for \texttt{Wholesaler2} where the \texttt{Exchange Rate} data set is the closest.

\subsection{Source data set diversity}
To measure the diversity per source data set, we consider the variance of the two feature sets calculated in Section \ref{sec:TS Features}. The variance of each feature is Min-Max scaled between zero and one over the source data sets. Hence, the source with the highest variance of a feature in comparison to the other sources, is indicated by a one for this feature. Then the sum over the feature variances is built and saved as diversity value for the respective source data set. 

\begin{table}[ht]
    \centering
    \begin{tabular}{l|rr}
    \hline
        Source & tsfresh & catch22 \\ \hline
        \texttt{Multisource} & 1.73 & 8.95 \\ 
        \texttt{M5} & 1.54 & 8.25 \\ 
        \texttt{M4} & 4.01 & \textbf{12.86} \\ 
        \texttt{Electricity} & \textbf{5.51} & 6.08 \\ 
        \texttt{NN5} & 0.69 & 5.15 \\ 
        \texttt{Exchange Rate} & 0.03 & 6.34 \\ \hline
    \end{tabular}
    \caption{Source data set diversity based on tsfresh and catch22 features. The diversity values of the most divese source data according to the features are printed in bold.}
    \label{source_diversities}
\end{table}

The diversity based on the tsfresh and catch22 features of the source data sets utilized are given in Table \ref{source_diversities}. According to the tsfresh features, the \texttt{Electricity} is the most diverse source and according to the catch22 features the \texttt{M4} data set.

\subsection{Visualization of data set diversity}
\label{sec: pca}
To visualize the data set diversity, we implement a two-dimensional principal component analysis (PCA) of the time series features of both sets. We standardize their features over all data sets, so that they get projected into the same latent space. We compute the PCA with two principal components via the \textit{sklearn} package~\citep{scikit-learn}.

The principal components' loading vectors of the tsfresh feature set can be found in the Appendix Table \ref{pca_loadings_tsfresh}. Here, we summarize what we conclude from it. Features related to the location and spreading dimensions, like "energy", "mean", "median", and "standard deviation" are most important for the first component, while shape-related features like the "skewness", "kurtosis", and "intermittency" are most important for the second. This means that these features drive the variance in the data sets and are predominantly responsible for the positions of the time series within the latent space spanned by the principal components. The variance explained by the PCA is 49\% and 30\% respectively for the two components and 79\% overall, indicating a meaningful analysis. 

\begin{figure}[ht]
  \centering
  \subfloat[\centering Sources Tsfresh]{{\includegraphics[width=7.5cm]{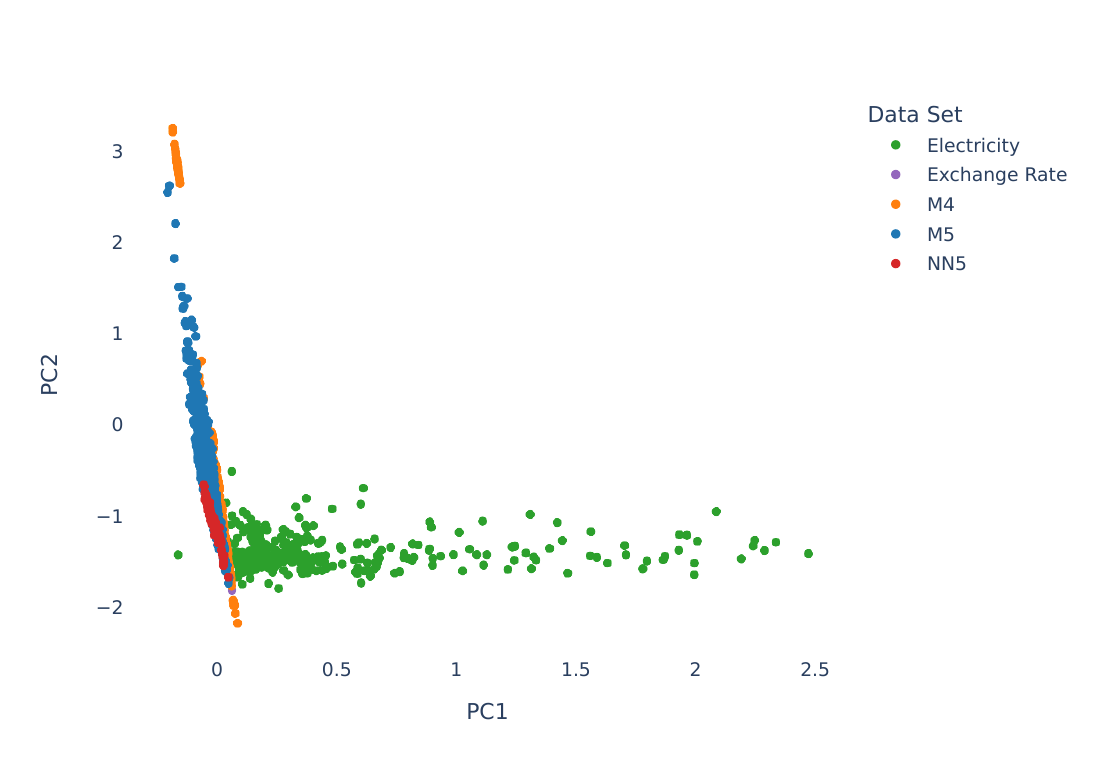}}}
  \qquad
  \subfloat[\centering Targets Tsfresh]{{\includegraphics[width=7.5cm]{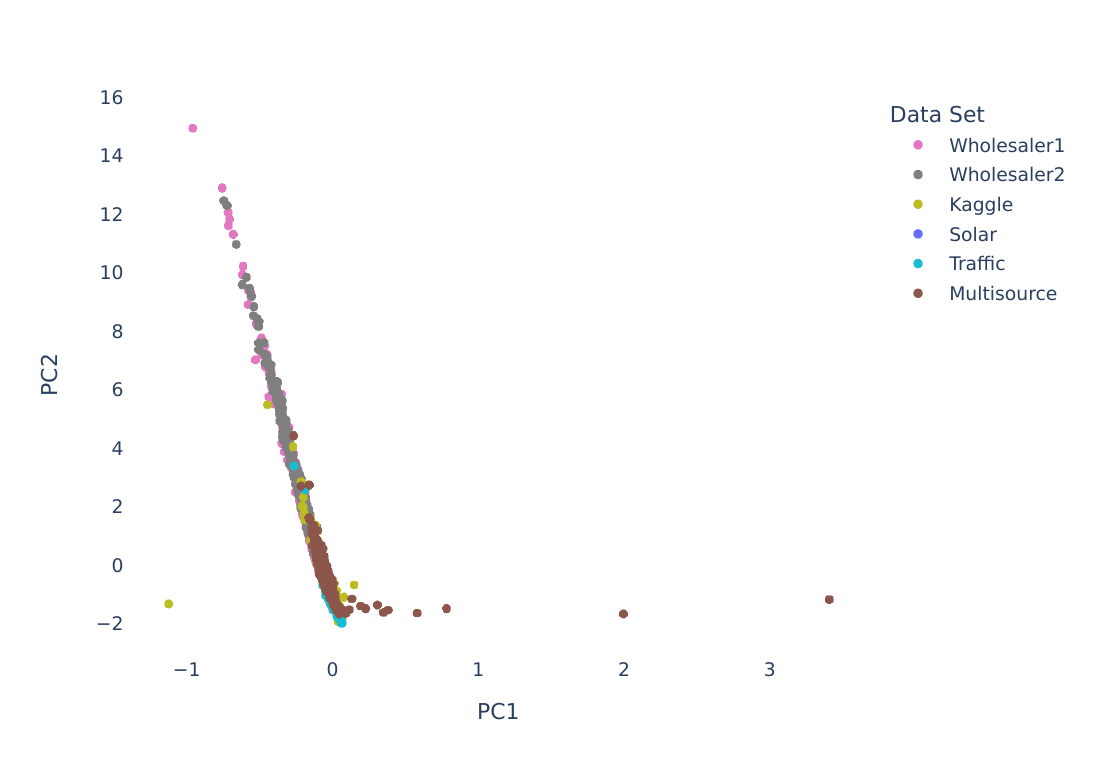}}}
  \qquad
  \subfloat[\centering Sources Catch22]{{\includegraphics[width=7.5cm]{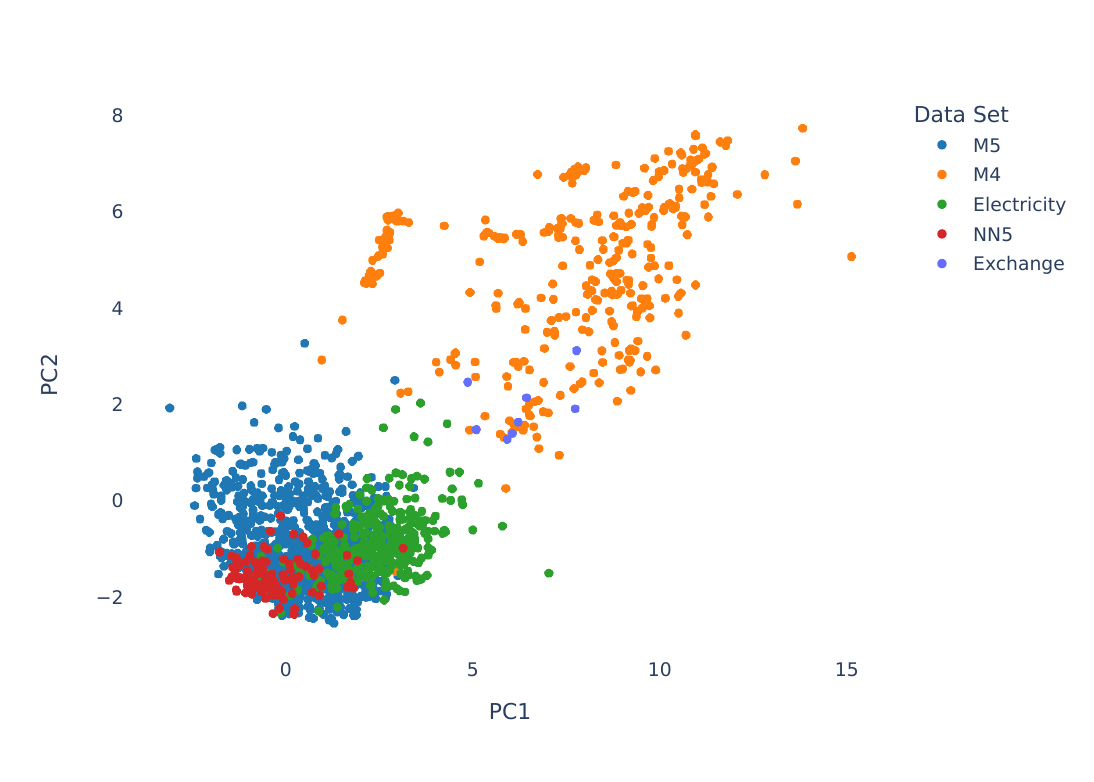}}}
  \qquad
  \subfloat[\centering Targets Catch22]{{\includegraphics[width=7.5cm]{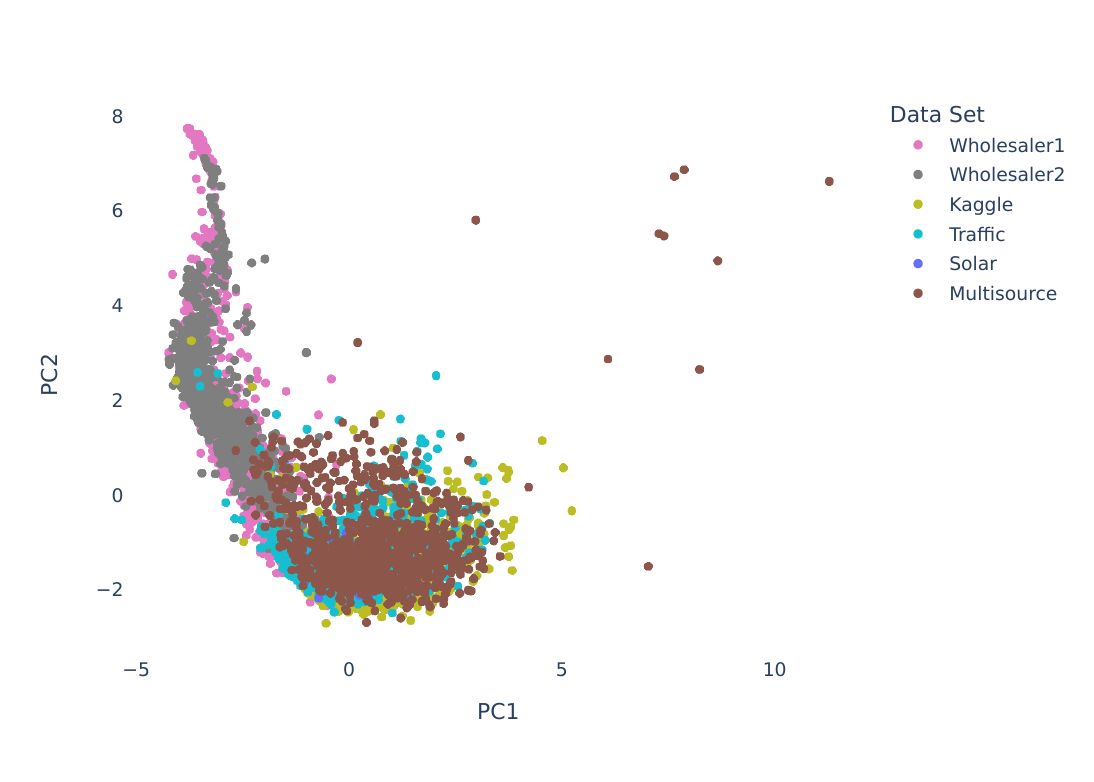}}}
  \caption{PCAs of the tsfresh and catch22 features of the source and target data sets with the latter including the \texttt{Multisource} one. Best viewed in color.}
  \label{pca}
\end{figure}

The PCA plot of the source data sets according to the tsfresh features can be seen in the upper left corner of Figure~\ref{pca}. The \texttt{Electricity} data set is the most diverse source, as seen in Table~\ref{source_diversities}, and also spreads more widely across the instance space than the other data sets. The y-axis spread of the \texttt{M4} data set is larger than that of the \texttt{M5}, which is in line with their diversity measures in Table~\ref{source_diversities}. 

The upper right image of Figure~\ref{pca} shows the \texttt{Multisource} data set together with the target data sets in the latent space according to the tsfresh features. The former has inherited its x-axis spread from the \texttt{Electricity} data set. The area around the origin is much more covered by the \texttt{Multisource} than the higher y-axis, where a considerable amount of target time series live. The two wholesales data sets have similar projections. The PCA plot suggests, that the sources cannot fully cover the targets' space. This might be due to their different domains.

The Appendix Table~\ref{pca_loadings_catch22} lists the loading vectors for the catch22 features. Here, these factors  explain 30\% and 19\% of the variance for each component and 49\% together, indicating a mediocre analysis. For the first component "ami2", "centroid frequency", "forecast error" are the most important ones and for the second "transition matrix", "entropy pairs", "high fluctuation". As opposed to the tsfresh features, the former are less interpretable.

The bottom left image of Figure~\ref{pca} shows the source data sets according to their catch22 features in the PCA space. The picture is completely different to that of the tsfresh features, as are the respective source diversity values in Table~\ref{source_diversities}. It demonstrates that the \texttt{M4} data set is quite different from the others in terms of location and spread, which is in line with being the most diverse according to Table~\ref{source_diversities}. The bottom right image of Figure~\ref{pca} visualizes the target data sets according to the catch22 features together with the \texttt{Multisource}. As for the tsfresh features, the wholesales data sets occupy the same space. Again, the time series of all sources, represented by the \texttt{Multisource}, do not cover all the space of the targets.

\section{Forecasting performance vs. source-target similarity and source diversity} 
\label{sec: relations_results}
Given the analysis of similarities between source and target data sets and of diversities of the sources, we now assess their relations to transfer learning forecasting performance on the targets. To that end, we create separate plots with each showing one link between one of the transfer learning accuracy (see Table~\ref{acc_results}), bias (see Table~\ref{bias_results}) and uncertainty estimation (see Table~\ref{uncertainty_results}) measures and one of the similarity by tsfresh feature distance (see Table~\ref{tsfresh_distances}), catch22 feature distance (see Table~\ref{catch22_distances}), DTW distance (see Table~\ref{dtw_distances}), source diversity by tsfresh, source diversity by catch22 features (for both see Table~\ref{source_diversities}) measures, for either zero-shot or fine-tuning forecasts for all target data sets. To test for linear relationships, we fit a linear regression to each target relation line and report the slope and its p-value.

In the following, we show only those plots together with their linear regression fitting information for which we find a roughly linear relationship throughout the targets. They only include zero-shot forecasting performance, since fine-tuning harmonizes forecasting performance, as seen in Section~\ref{sec:exp1_results}, which consequently flattens the trends in their relation to similarity and diversity values. This suggests that fine-tuning makes forecasts more independent of the source data set used. 

\subsection{The effect of source diversity}
\begin{figure}[!ht]
    \centering
    \subfloat[\centering Catch22 diversity and zero-shot forecast accuracy error]{{\includegraphics[width=7.5cm]{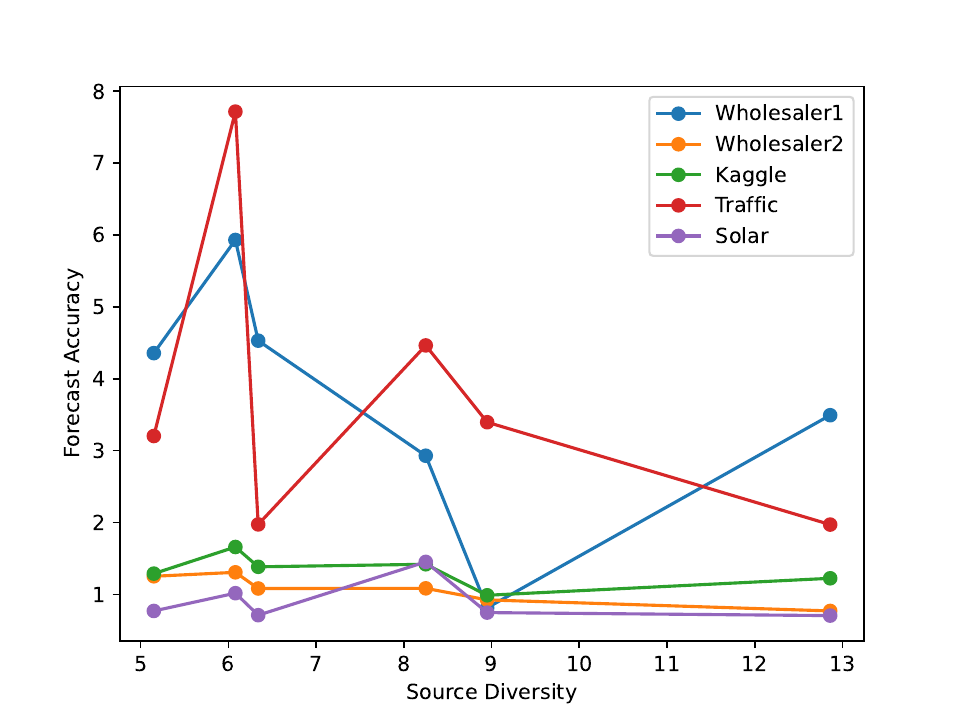}}}
    \qquad
    \subfloat[\centering Tsfresh diversity and zero-shot bias]{{\includegraphics[width=7.5cm]{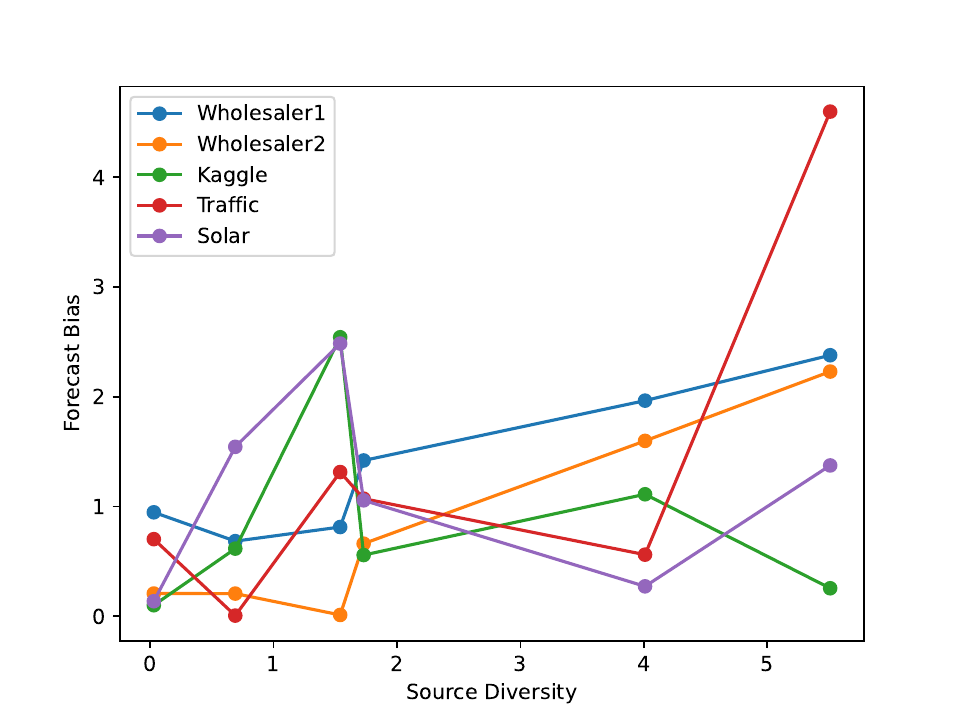}}}
    \qquad
    \subfloat[\centering Catch22 diversity and zero-shot uncertainty estimation]{{\includegraphics[width=7.5cm]{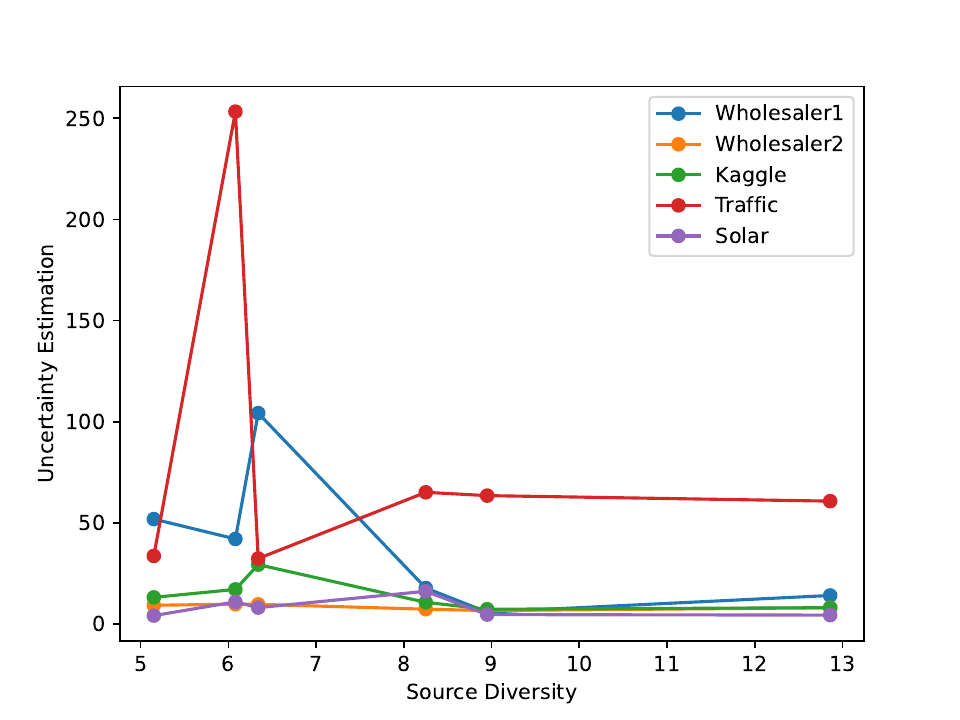}}}
    \caption{Relation between the catch22 (a+c) and tsfresh (b) source diversity and their zero-shot transfer learning forecast accuracy error (a), bias (b) and uncertainty estimation error (c). Each curve corresponds to a target data set and each point to a source data set with its respective diversity on the x-axis and its zero-shot transfer learning forecasting performance on the y-axis.}
    \label{diversity}
\end{figure}

\begin{table}[ht]
    \centering
    \begin{tabular}{l|ll|ll|ll}
    \hline
        Target & Acc\_slope & P\_val & Unc\_slope & P\_val & Bias\_slope & P\_val\\ \hline
        Wholesaler1 & -0.29 & 0.34 & -7.88 &  0.20  & \textbf{0.31} & 0.01  \\ 
        Wholesaler2 & \textbf{-0.07} & 0.01 & -0.28 & 0.21 & \textbf{0.41} & 0.00   \\
        Kaggle & -0.04 & 0.36 & -1.66 & 0.24 & -1.66 & 0.24  \\ 
        Traffic & -0.30 & 0.44 & -5.95 & 0.71 &  -5.95 & 0.71  \\ 
        Solar & -0.01 & 0.82 & -0.34 & 0.70 & -0.34 & 0.70  \\ \hline
    \end{tabular}
    \caption{Linear regression fit of catch22 diversity and zero-shot forecast accuracy and uncertainty error, and tsfresh diversity and zero-shot forecast bias, per target data set. Statistically significant slope values are printed in bold.}
    \label{lr_diversity}
\end{table}

Figure~\ref{diversity}(a) and (c) demonstrate that more diversity in catch22 features is favorable for zero-shot accuracy and uncertainty estimation. This can be seen by the target curves going from higher forecast accuracy errors and uncertainty estimation errors to lower ones with increasing source diversity. \texttt{Traffic} and \texttt{Wholesaler1} provide a broader range of error values. The \texttt{Solar} and \texttt{Kaggle Wiki} data sets show more of a flat progression, but do not show the opposite of the other curves. 
When fitting a linear regression per target data set, all slopes are negative in the columns of Table~\ref{lr_diversity}, which correspond to accuracy and uncertainty, indicating that more diversity improves forecast accuracy and uncertainty estimation. Note that for the accuracy, only for \texttt{Wholesaler2} the p-values of the slope are statistically significant. For the uncertainty none of the slope coefficients are significant.

Further, Subfigure (b) shows that sources with a higher diversity in tsfresh features produce transfer learning forecasts with higher bias. A look at the underlying Tables~\ref{source_diversities} and~\ref{bias_results} shows that e.g. the \texttt{Electricity} source data set, which has the highest tsfresh diversity, shows the highest zero-shot transfer learning forecast biases over the targets. The linear regression curves show almost always a positive slope (bias columns in Table~\ref{lr_diversity}), which are significant for the wholesalers. 

\subsection{The effect of source-target similarity}
\begin{figure}[ht]
    \centering  
    \includegraphics[width=7.5cm]{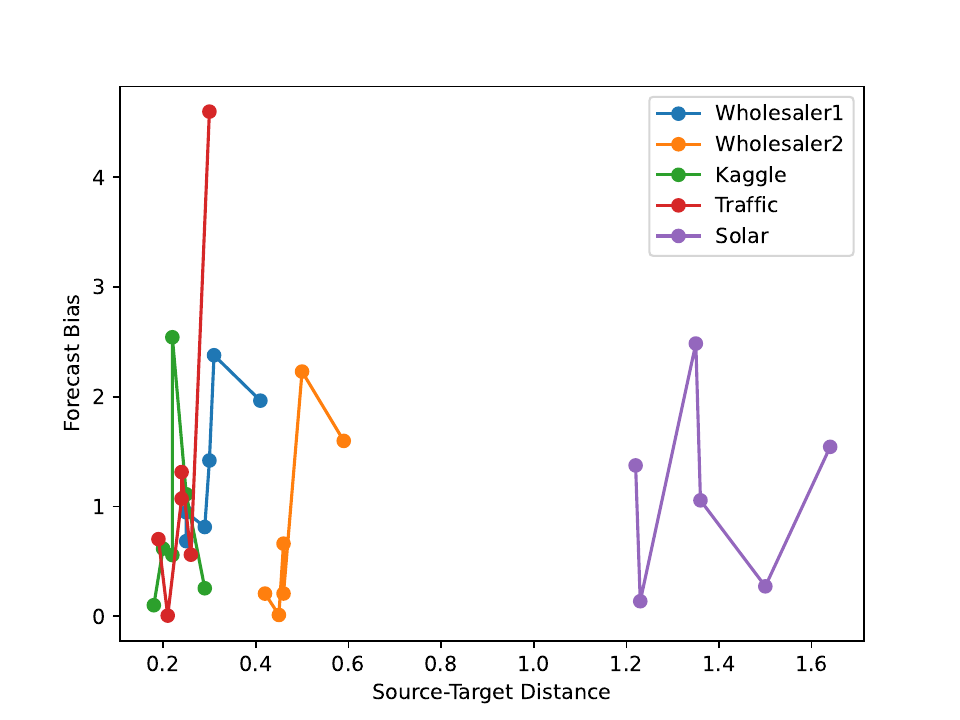}
    \caption{Relations between tsfresh source-target distance and zero-shot transfer learning bias. Each curve corresponds to a target data set and each point to a source data set with its respective tsfresh feature distance on the x-axis and the zero-shot transfer learning bias on the y-axis.}
    \label{similarity}
\end{figure}

\begin{table}[ht]
    \centering
    \begin{tabular}{l|ll}
    \hline
        Target & Bias\_slope & P\_val \\ \hline
        Wholesaler1 & 7.90 & 0.14  \\ 
        Wholesaler2 & 10.69 & 0.11 \\ 
        Kaggle & 0.04 & 1.00 \\ 
        Traffic & 34.04 & 0.06 \\ 
        Solar & 0.57 & 0.84 \\ \hline
    \end{tabular}
    \caption{Linear regression fit of tsfresh distance and zero-shot forecast bias per target data set.}
    \label{lr_similarity_bias}
\end{table}

Figure~\ref{similarity} shows that source data sets that are more similar to the target data sets in their tsfresh features create less forecast bias. This can be seen by the target curves going from lower forecast bias to higher ones as the distance to the sources increases. All linear regression curves provide a positive slope (Table~\ref{lr_similarity_bias}), which is not statistically significant.

Note that the relations are relative for the source data sets of each target, as we scale over the source data sets to get similarity and diversity measures. Hence, the relations are not absolute, which can be seen at the \texttt{Solar} data set for example, which has the largest distance to the source data sets but does not provide the worst forecast performance.

We could not find such consistent relations regarding the accuracy, uncertainty estimation, catch22 features or DTW distances. 

\section{Conclusion} 
\label{sec:conclusion}
To conclude, we provide a summary of the methods and results, as well as limitations and directions for further research.
\subsection{Summary}
We investigated the impact of source-target data set similarity and source data set diversity onto the success of transfer learning. Therefore, we assessed zero-shot and fine-tuned forecasting performance regarding accuracy, bias and uncertainty estimation of five pre-trained single-source models, their ensemble and the foundation model trained on the combination of the source data sets, on five diverse real-world target data sets. After that, we analyzed the relations of source and target data sets by calculating two feature-based and one shape-based similarity measures. The diversity of the source data sets was computed as the variance of the two feature sets and visualized by projecting the data sets into the latent PCA space.

We find that high similarity between the source and target data set in tsfresh features enhances the zero-shot bias. A high diversity of the source data set in catch22 features leads to better zero-shot accuracy and uncertainty estimation, while high diversity in tsfresh features increases zero-shot bias.
As opposed to the feature-based similarity measures, the shape-based DTW distance measure does not show any relation to the forecasting results.
Fine-tuning smoothens the relations between similarity, diversity and forecasting performance. Hence, it is more relevant to investigate data set characteristics in regard to transfer learning success in the zero-shot case. If there are resources, we recommend to fine-tune pre-trained models for enhanced transfer learning accuracy and uncertainty estimation; then the source data set does not play a major role, according to our findings. 
As stated above, feature-based similarity and diversity measures reveal different relations to the forecasting performance and are therefore both helpful in choosing a source data set out of a variety. According to our findings, we recommend choosing the source data set which is either the most similar in tsfresh features or the most diverse in catch22 features. For the latter, our pre-trained model on the \texttt{M4} data can be used.
Our results justify the approaches in the literature of selecting similar source data sets for a specific target or building a foundation model with diverse source data sets.
Regardless of the source domain, similarities in tsfresh features can be an indicator of zero-shot transfer learning success.

\subsection{Limitations and research directions}
The source data sets considered in our study were not chosen to be similar to any of the target data sets or to be as diverse as possible. Hence, our results rely on random source data sets, of which we can measure their similarity to the targets and their diversity only in a relative manner, while the closest source in our experiment might still be not similar in an absolute sense and the most diverse source may not be diverse in comparison to much larger data sets used for training foundation models. Therefore, we can only supply relative results and no similarity or diversity metric with absolute thresholds telling whether a source should be used or not to guarantee for transfer learning success.

Many of the results shown in Section~\ref{sec: relations_results}, especially the p-values of the regression lines, are not statistically significant. A reason might be the number of data points. Hence, we call for further studies examining these relationships with more source and target data sets.

Further, the models and results presented here are restricted to weekly time series data. Including multiple granularities and also transfer learning across them is outside the scope of this paper. We also exclude multivariate time series or external regressors, since the goal is to identify relations between similarities and diversities to forecasting performance and not relations between time series. Nevertheless, it would be interesting to see how the additional information, which is contained in a couple of data sets of the Monash Forecasting Repository~\citep{Godahewa.2021}, changes the relations. Furthermore, we use the same forecast horizon of fifteen weeks for all data sets for training and testing. Nevertheless, the pre-trained DeepAR models applied and published (upon acceptance) for transfer learning are able to do zero-shot and fine-tuned forecasts also for other horizons. The results might be different for other horizons. We leave these topics for further research.

From a broader perspective, comparing general foundation models, which are trained on data sets of diverse domains, to domain specifically pre-trained models might be an interesting task. This is indicated in this study by the \texttt{M5} data set used as source performs better on the wholesales data sets than on the other targets. Using other source tasks like imputing missing data might also be beneficial for pre-training models for time series forecasting.

\section*{Acknowledgements}
Special thanks go to  Leif Feddersen, Ali Ismail-Fawaz and Julius Mehringer for thoughtful discussions on the ideas around this paper. Also we would like to thank Julia Schemm, Nico Beck and Vrinda Gupta who co-setup the basic code on which the experiments were built upon. We extend our sincere gratitude to the providers of the open datasets, whose generous sharing of data resources has been instrumental in the advancement of our research.

\bibliography{refs}

\section*{Appendix}
\subsection*{Training time}
\begin{table}[ht]
    \centering
    \begin{tabular}{r|rrrrrr}
        Source / Target & \texttt{Solar} & \texttt{Traffic} & \texttt{Kaggle Wiki} & \texttt{Wholesaler2} & \texttt{Wholesaler1} & Sum\\ \hline
        \texttt{Multisource} & \textbf{0.32} & 2.17 & 0.37 & 0.37 & \textbf{0.06} & 3.29\\ 
        \texttt{Ensemble} & 2.34 & 2.70 & 3.95 & 3.98 & 0.42 & 13.39 \\ 
        \texttt{M5} & 0.55 & 0.65 & \textbf{0.22} & 1.98 & \textbf{0.06} & 3.46\\ 
        \texttt{M4} & \textbf{0.32} & 1.03 & \textbf{0.22} & \textbf{0.25} & \textbf{0.06} & 1.88\\ 
        \texttt{Electricity} & 0.52 & 0.43 & 2.17 & 1.02 & \textbf{0.06} & 4.20 \\ 
        \texttt{NN5} & 0.41 & 0.33 & 1.08 & 0.37 & 0.16 & 2.35\\ 
        \texttt{Exchange Rate} & 0.54 & \textbf{0.27} & 0.26 & 0.35 & 0.08 & \textbf{1.50}\\ 
    \end{tabular}
    \caption{Relative fine-tuning times on the target data sets of the pre-trained models, trained on the respective source data sets, relative to training from scratch. Values smaller than one indicate faster fine-tuning than training from scratch. The smallest fine-tuning times per target are printed in bold.}
    \label{times}
\end{table}

Figure \ref{times} shows the relative training times of fine-tuning a pre-trained model on the target data compared to training from scratch. Since both is done in an early stopping manner, it is comparable. Hence, fine-tuning times smaller than 1 mean that it is faster to fine-tune the respective pre-trained model than training from scratch i.e. one can save training time by using a pre-trained model. This is the case in 71\% of the time. The \texttt{Ensemble} model takes the most time to fine-tune, since each ensemble member, which corresponds to a pre-trained model on one of the source data sets, needs to be fine-tuned on the target data. For the target data set \texttt{Wholesaler1}, which is the largest in this study, it is faster to fine-tune any pre-trained model than training from scratch. 

In four of five cases, it is fastest to fine-tune the model pre-trained with the \texttt{M4} data, which already provided a good zero-shot performance in Table~\ref{acc_results}. Summed over all target data sets, the model pre-trained on the \texttt{Exchange Rate} data, which is the smallest source data set, takes the least time to fine-tune. Without the \texttt{Ensemble} model, the pre-trained \texttt{Electricity} model takes the longest to fine-tune. A possible reason might be the bad zero-shot performance of the latter seen in Table~\ref{acc_results}, far away from an optimum of the model's parameters for each target.

To sum up, in most cases it saves time to fine-tune a pre-trained model instead of training it from scratch. This is one advantage of transfer learning, which is also shown in literature, see Section~\ref{sec:related_work}.

\subsection*{Tables}
\begin{table}[ht]
\centering
\hspace*{-1cm}
    \begin{tabular}{|l|p{7cm}|l|}
    \hline
        Name & Explanation & Source \\ \hline
        Absolute Energy & Sum of squared values & tsfresh \\ \hline
        Intermittency & Length of time series divided \newline by the sum of non-zero time steps & own \\ \hline
        Mean & Mean of the time series & tsfresh \\ \hline
        Median & Median of the time series & tsfresh \\ \hline
        Kurtotsis & Kurtosis of the time series & tsfresh \\ \hline
        Skewness & Skewness of the time series & tsfresh \\ \hline
        Standard deviation & Standard deviation of the time series & tsfresh \\ \hline
        Aggregate autocorrelation max & Maximal autocorrelation of the first five lags of the time series & tsfresh \\ \hline
        Lumpyness & Variance of non-zero time steps divided \newline by the squared mean of non zero time steps & own \\ \hline
        Linear Trend & Slope of the time series & tsfresh \\ \hline
    \end{tabular}
\caption{Tsfresh features, also including two of our own.}
\label{tsfresh_feat}
\end{table}

\begin{table}[ht]
    \centering
    \begin{tabular}{|r|r|r|r|r|}
    \hline
        energy & intermittency & mean & median & kurtosis  \\ \hline
        0.45 & -0.02 & 0.45 & 0.44 & -0.03  \\ \hline
        0.04 & 0.50 & 0.02 & 0.02 & 0.54  \\ \hline \hline
        skewness & sd & auto corr & lumpyness & trend \\ \hline
        -0.03 & 0.45 & 0.04 & -0.02 & -0.45 \\ \hline
        0.55 & 0.03 & -0.31 & 0.23 & -0.03\\ \hline
    \end{tabular}
    \caption{Loading vectors of the principal components of the tsfresh features.}
    \label{pca_loadings_tsfresh}
\end{table}

\begin{table}[ht]
    \centering
    \begin{tabular}{|r|r|r|r|r|}
    \hline
        mode\_5 & mode\_10 & acf\_timescale & acf\_first\_min & ami2\\ \hline
        -0.12 & -0.14 & 0.31 & 0.21 & 0.33   \\ \hline
        0.31 & 0.22 & 0.08 & 0.2 & -0.07  \\ \hline \hline
        trev & high\_fluctuation & stretch\_high & transition\_matrix & periodicity \\ \hline
        0.04 & -0.11 & 0.3 & -0.04 & 0.23 \\ \hline
        -0.25 & -0.4 & 0.07 & 0.4 & 0.07 \\ \hline \hline 
        embedding\_dist & ami\_timescale & whiten\_timescale & outlier\_timing\_pos & outlier\_timing\_neg \\ \hline
        0.26 & 0.26 & -0.18 & -0.12 & 0.14 \\ \hline
        -0.03 & 0.21 & 0.29 & -0.18 & 0.08 \\ \hline \hline
        centroid\_freq & stretch\_decreasing & entropy\_pairs & rs\_range & dfa  \\ \hline
        0.31 & 0.27 & -0.08 & 0.0 & 0.08 \\ \hline
        -0.11 & -0.11 & -0.43 & 0.01 & 0.02 \\ \hline \hline
        low\_freq\_power & forecast\_error & mean & sd & \\ \hline
        -0.27 & -0.33 & 0.02 & 0.01 & \\ \hline
        0.18 & -0.06 & -0.05 & -0.03 & \\ \hline
    \end{tabular}
    \caption{Loading vectors of the principal components of the catch22 features.}
    \label{pca_loadings_catch22}
\end{table}

\end{document}